\documentclass{article}

\usepackage[final]{neurips_2023}

\usepackage[utf8]{inputenc} % allow utf-8 input
\usepackage[T1]{fontenc}    % use 8-bit T1 fonts
\usepackage{url}            % simple URL typesetting
\usepackage{booktabs}       % professional-quality tables
\usepackage{amsfonts}       % blackboard math symbols
\usepackage{nicefrac}       % compact symbols for 1/2, etc.
\usepackage{microtype}      % microtypography
\usepackage{xcolor}         % colors
\usepackage{tabularray}
\usepackage{multirow}
\usepackage{hyperref}
\usepackage{amsmath}
\usepackage[ruled,vlined]{algorithm2e}
\usepackage{tablefootnote}

\DeclareMathOperator*{\argmin}{arg\,min}

\definecolor{airforceblue}{rgb}{0.36, 0.54, 0.66}

\usepackage{amsmath,bm}
\usepackage{subcaption}
\usepackage{graphicx}
\usepackage{stfloats}
\newcommand{\rvline}{\hspace*{-\arraycolsep}\vline\hspace*{-\arraycolsep}}

\title{A supervised generative optimization approach for tabular data}

\author{%
Shinpei Nakamura-Sakai \footnotemark[1] \\
Department of Statistics and Data Science\\
Yale University \\
New Haven, Connecticut USA \\
\texttt{s.nakamura.sakai@yale.edu} \\
\AND
  Fadi Hamad \thanks{Both authors contributed equally to this research and order is exchangeable}\\
  Department of Industrial Engineering\\
  University of Pittsburgh\\
  Pittsburgh, PA USA \\
  \texttt{fah33@pitt.edu} \\
   \And
  Saheed Obitayo \\
  J.P. Morgan AI Research \\
  New York, USA \\
  \texttt{saheed.o.obitayo@jpmorgan.com} \\
  \And
  Vamsi K. Potluru \\
  J.P. Morgan AI Research \\
  New York, USA \\
  \texttt{vamsi.k.potluru@jpmchase.com} \\
}

\begin{document}

\maketitle

\begin{abstract}
  Synthetic data generation has emerged as a crucial topic for financial institutions, driven by multiple factors, such as privacy protection and data augmentation. Many algorithms have been proposed for synthetic data generation but reaching the consensus on which method we should use for the specific data sets and use cases remains challenging. Moreover, the majority of existing approaches are ``unsupervised'' in the sense that they do not take into account the downstream task. To address these issues, this work presents a novel synthetic data generation framework. The framework integrates a supervised component tailored to the specific downstream task and employs a meta-learning approach to learn the optimal mixture distribution of existing synthetic distributions.
\end{abstract}

\maketitle

% \tableofcontents

% CTabGAN time: 1277.005440711975  seconds+0.45 for fitting adult AUC= 0.8571508423115641
% 625.373363494873+0.28 for bcredit AUC=0.9730104817461139
% 621.0299413204193+0.29 for ucredit AUC=0.9460141167767722

%, data,AUC
%0,adult,0.8683547705656879
%1,encoded_adult,0.8357537995969538
%2,credit_card_balanced50k,0.9809648328988659
%3,credit_card_unbalanced50k,0.9266574883644472

\section{Introduction}
Synthetic data generation is vital in various industries, like finance, telecommunication, and healthcare where data-driven decision-making is crucial \cite{jordon2022synthetic}. It resolves data scarcity and quality concerns by providing synthetic data that preserves statistical properties and relations with the original data. Synthetic data enables testing new ideas without compromising real data, blending multiple sources, and protecting individual privacy \cite{voigt2017eu}. However, using synthetic data may cause performance degradation in modeling \cite{hittmeir2019utility} where utility degradation depends on the fidelity of the data generation process and the downstream task. To address this issue and maintain synthetic data quality, the development of a framework to mitigate the degradation is indispensable.

The majority of existing approaches are ``unsupervised'' in the sense that they do not take into account the downstream task. For instance, the methods discussed in \cite{patki2016synthetic} treat the output variable used in the downstream task like other covariates. Their primary focus is creating models that merely ``resemble'' the original data sets, with variations in how they quantify this resemblance. However, in some use cases, the primary objective is not just achieving similarity to the actual data distribution but optimizing for downstream predictions using the generated data. While research suggests \cite{hittmeir2019utility} that a closer match to the original data distribution generally leads to better downstream performance, there is an alternative approach worth exploring, incorporating a ``supervised'' component and optimizing directly on the downstream loss function. By doing so, the generated data can be tailored more effectively to improve the performance of the specific downstream task.

Recent studies show that there are various approaches to model tabular data distribution and sample rows from the distribution\cite{eno2008generating}: neural network-based\cite{park2018data}, machine learning-based\cite{caiola2010random}, and statistical-based\cite{li2020sync} generative models. Each of these synthetic data generation methods has its distinct capabilities and features. Many algorithms have been proposed for synthetic data generation but reaching the consensus on which method we should use for the specific data sets and use cases remains challenging. For instance, \cite{xu2019modeling} found that, while table variational autoEncoder (TVAE) generally outperforms Conditional Tabular Generative Adversarial Networks (CTGAN) in most experiments, CTGANs are preferred when prioritizing differential privacy. Furthermore, Gaussian Copula, being a parametric model, proves advantageous when dealing with known marginal distributions. More importantly, the appropriate method depends on many factors such as the distribution of the observed data or the objective of generating the synthetic data. 

This work proposes a novel synthetic data generation framework, the Supervised and Composed Generative Optimization Approach for Tabular Data (SC-GOAT), to address the aforementioned issues. The framework comprises two key steps. Firstly, we incorporate a supervised component customized for the specific downstream task. To achieve this, we leverage a Bayesian optimization approach to fine-tune the hyperparameters related to the neural networks. Additionally, apart from optimizing the traditional loss of GAN or TVAE through their parameters, we also tune the hyperparameters based on a loss function specific to the downstream task on the validation set. Subsequently, we evaluate the model's performance based on the test error to accurately assess its effectiveness.

In the second step, we adopt a meta-learning approach, leveraging Bayesian optimization, to identify the optimal mixture distribution of existing synthetic data generation methods. To the best of our knowledge, this approach is the first to generate synthetic data based on a mixture of multiple synthetic data generation methods. From each method we learned in the first step, we explored multiple data generation techniques and tuned the proportion of data sets sampled. This approach is motivated by the quest to discover the projection of the true underlying data distribution onto the set encompassing various synthesizers. 

Employing supervised Bayesian optimization, we search for the ideal mixture that optimizes the downstream performance metric. By dynamically combining the strengths of different data generation methods, we aim to enhance the overall synthetic data quality and its suitability for downstream tasks.

\textbf{Our contributions}
%Shinpei
\begin{enumerate}
    \item Our approach incorporates supervised components, granting us the flexibility to customize the metric of interest. Whether it's efficacy, fidelity, or privacy, we can tailor the approach accordingly to boost its performance. 
    
    \item  %To take advantage of the different types of existing methods, 
    We introduce a meta-learning framework that leverages various methods to learn the optimal mixture distribution, improving our metric of interest. Additionally, our approach remains robust, even with inefficient synthetic data from certain models.
    \item Our proposed methodology consistently outperforms existing methods, exhibiting a statistically significant improvement with a p-value of less than 1\% in the majority of cases.
\end{enumerate}

\section{Related Work}\label{sec:related-word}
Recent studies have revealed diverse approaches to modeling tabular data distribution and sampling from it \cite{eno2008generating}. These approaches include Neural network-based methods \cite{park2018data}, Machine Learning-based techniques \cite{caiola2010random}, and Statistical-based generative models \cite{li2020sync}. Each of these methods for synthetic data generation possesses unique capabilities and features. For the purpose of this paper, our focus centers on exploring Neural network-based and Statistical-based approaches.
% Recent studies show that there are various approaches to model tabular data distribution and sample rows from the distribution\cite{eno2008generating}: Neural network-based\cite{park2018data}, Machine Learning-based\cite{caiola2010random}, and Statistical-based\cite{li2020sync} generative models. Each of these synthetic data generation methods has its distinct capabilities and features. In this paper, we focus our attention on Neural network-based and statistical-based approaches. 

The Synthetic Data Vault (SDV) project, utilized for conducting most of the experiments \cite{patki2016synthetic}, offers two Generative Adversarial Networks (GAN)-based models for data generation from single tables: Conditional Tabular GAN (CTGAN) and CopulaGAN. GANs represent a powerful generative modeling approach employing deep learning methods like convolutional neural networks. Since the original GAN formulation \cite{goodfellow2014generative}, ongoing research has led to the proposal of new optimization strategies and modifications to address GAN limitations. One notable model that builds upon prior successes is CTGAN, which employs mode-specific normalization to capture non-Gaussian and multimodal distributions \cite{xu2019modeling}. It also introduces a conditional generator and training by sampling to tackle challenges posed by highly imbalanced categorical columns and the sparsity of one-hot-encoded vectors, limitations observed in previous GAN architectures. Another neural network approach from SDV is known as TVAE \cite{xu2019modeling}, which is the variational encoders adapted for tabular data. 

Beyond neural networks, synthetic data generation can also be achieved by treating each table column as a random variable, modeling a multivariate probability distribution, and sampling from it. SDV presents a synthesizer using this approach called Gaussian Copula \cite{masarotto2012gaussian}, which leverages copula functions. These mathematical functions allow for describing the joint distribution of multiple random variables by analyzing the dependencies between their marginal distributions \cite{patki2016synthetic}. In the SDV project, univariate marginals are learned using a Gaussian mixture model, while the multivariate copula is learned as a Gaussian copula. The Gaussian Copula approach is valuable for modeling both the covariances between features and their distributions \cite{llugiqi2022empirical}.

On the other hand, Bayesian optimization is a powerful and efficient technique used in various fields to optimize complex and costly functions \cite{pelikan1999boa,snoek2012practical}. This methodology is particularly valuable when exploring black-box functions, where the underlying mathematical form is unknown or computationally expensive to evaluate. Additionally, Bayesian optimization is used when the hyper-parameters space isn't continuous or the loss function isn't differentiable. At its core, Bayesian optimization employs a probabilistic model, typically a Gaussian Process, to capture the surrogate representation of the objective function. By iteratively selecting the next sampling point based on a trade-off between exploration and exploitation, it intelligently navigates the search space, efficiently narrowing down the region likely to contain the global optimum \cite{frazier2018tutorial}. This approach has shown remarkable success in tasks like hyperparameter tuning and parameter optimization in machine learning, engineering \cite{snoek2012practical, frazier2018tutorial}, and other domains.

\section{Synthetic Data Generation}\label{sec:our-approach}

Let $M = \{$GC,  CTGAN, C-GAN, TVAE$\}$ be the set of the synthetic data generation methods being utilized. For each method $m \in M$, we have a corresponding synthetic data generation function $S_m(N;\omega_m;\theta_m)$ where $N$ is the number of rows to simulate, $\omega_m$ is the set of parameters, and $\theta_m$ is the set of hyper-parameters. Note that $\theta_{GC}=\emptyset$ as Gaussian Copula does not use neural networks.
 
Let $D_{real}$ represent the real data set and $D_m$ denote the synthetic data generated by model $m \in M$. Additionally, we have three data sets: $D_{train}$, $D_{val}$, and $D_{test}$, representing the training, validation, and testing data sets, respectively. All $D_{*}$ has an outcome vector and covariate matrix which could be represented as duplet $D_{*}=(X_{*}, Y_{*})$. The downstream loss function is defined as $\mathcal{L}(Y, \hat{Y})$ where $\hat{Y}$ is the outcome predicted by the downstream prediction function $\mu=f(Y\sim X)$. where $f(Y\sim X)$ is the notation for a regression estimator but $f$ can be any machine learning estimator and $\mu$ denotes the learned function. Additionally, when $\mu$ is learned from the synthetic data generated by $S_m(N;\omega_m;\theta_m)$ we denote it as $\mu_{\omega(\theta_m)}=f(Y_m,X_m)$ where $(X_m, Y_m) = S_m(N;\omega_m;\theta_m)$.

SC-GOAT consists of two steps, supervising (Algorithm ~\ref{alg:S-GOAT}) and composing (Algorithm ~\ref{alg:C-GOAT}). In both steps of SC-GOAT, our approach follows standard optimization procedures by optimizing a loss function. However, unlike traditional methods, we adopt a Bayesian approach that constructs a probabilistic model around the involved parameters. Subsequently, we update these parameters based on the evaluation performance of the loss function. To establish the prior/posterior distribution over the objective function, we employ the Parzen-Tree Estimator. This allows us to effectively locate the parameter space's optimal region, maximizing the expected improvement in the loss function. By employing Bayesian optimization in this manner, we can efficiently fine-tune the synthesizer models and enhance the overall performance in generating data that closely resembles the real data set boosting the downstream performance.

\subsection{Supervising Synthesizers}\label{sec:supervising-synthesizers}
% Supervising synthesizers (Shinpei)
The first step of SC-GOAT involves tuning the hyperparameters using an optimization approach that is supervised by the downstream performance metrics. The optimization formulation is given in \eqref{eq:hyper-parameter-outer-optimization} and \eqref{eq:hyper-parameter-inner-optimization} as a bi-level optimization problem. To solve this hyper-parameter tuning optimization problem, we employ a Bayesian optimization approach \cite{franceschi2018bilevel}. The flexibility of Bayesian hyperparameter tuning allows for easy switching of the target function to optimize. Moreover, we have the option to incorporate privacy or fidelity regularization in addition to the downstream task. The pseudo-code for hyper-parameter tuning the given model $m$ is presented in Algorithm \ref{alg:S-GOAT}.

The supervising synthesizer optimization problem using bi-level formulation is given by:
\begin{flalign}
    \begin{aligned}
    \theta ^ * = \argmin_{\theta} \text{\quad} & \mathcal{L}(Y_{val}, \hat{Y}_{val}) \\
    \text{s.t \quad} & \hat{Y}_{val} := \hat{\mu}_{\omega ^* (\theta)}(X_{val})
    \end{aligned}
    \label{eq:hyper-parameter-outer-optimization}
 \end{flalign}

 st
 
\begin{flalign}
    \begin{aligned}
    \omega ^ * (\theta) = \argmin_{\omega} \mathcal{F}(\omega, \theta, D_{train})
    \end{aligned}
    \label{eq:hyper-parameter-inner-optimization}
 \end{flalign}

where the outer optimization problem \eqref{eq:hyper-parameter-outer-optimization} is minimizing the loss function on the validation set $D_{val}$ and the inner optimization problem \eqref{eq:hyper-parameter-inner-optimization} is minimizing the loss function denoted by $\mathcal{F}$ on the training set $D_{train}$ for synthesizer model $S$. Note that these functions, $\mathcal{L}$ and $\mathcal{F}$, are not necessarily the same and may be measured on different models. For instance, $\mathcal{F}$ always refers to the loss function used during the training of the synthesizer model $S$, whereas $\mathcal{L}$ refers to the model's performance on the validation set, possibly employing a different evaluation metric. Alternatively, $\mathcal{L}$ could also refer to the loss function for the downstream task performed by model $f$, as is the case in our approach.

\begin{algorithm} [tbh]
\caption{\textbf{S}upervising Step - \textbf{G}enerative \textbf{O}ptimization \textbf{A}pproach for \textbf{T}abular data (S-GOAT)}\label{alg:S-GOAT}
\textbf{Input:} $D_{real}=(X_{real}, Y_{real})$, N\\
Create partition $\{D_{train}, D_{val}, D_{test}\}=\{(X_{train}, Y_{train}),(X_{val}, Y_{val}), (X_{test}, Y_{test})$\} \\
Initialize $\hat{\theta}^0_m$ \\
\For{$k = 1, \dots, K$}{
    Fit $S_m$ using $D_{train}$\\
    Generate $D^k_m=(X^k_m,Y^k_m)$ where $(X_m^k,Y_m^k)=S(N;\omega_m;\theta_m^k)$\\
    Train $\hat{\mu}^k_{\theta_m^k}=f(Y^k_m\sim X^k_m)$ \\
    Compute $\hat{Y}_{val}^k=\hat{\mu}^k(X_{val})$\\
    Compute $l^k=\mathcal{L}(\hat{Y}_{val}^k,Y_{val})$\\
    Suggest $\hat{\theta}^{k+1}$ using Bayesian Optimization based on $\{\theta^0_m,...\theta^k_m\}$ and $\{l^0,...,l^k\}$\\

}
\textbf{return} $\theta_m^{k^*}$ where $k^*=\argmin_{k} l^k$\\
\end{algorithm}

% Fadi
\subsection{Composing Synthesizers}

The second step of SC-GOAT is the composing process. Here, our objective is to utilize a meta-learning approach to determine the mixture distribution among the synthesizers in $M$. We refer to it as a meta-learning approach because we learn the final model from the models obtained in the previous step. For each synthesizer $m\in M$, we define $\alpha_m\in [0,1]$ as the proportion of the total observations sampled from $S_m$. The final synthetic data comprises $[\alpha_m N]$ observations for each $m$, where $[\cdot]$ denotes the closest integer function. The formulation of this meta-learning approach using an optimization framework is given in \eqref{eq:meta-learning-optimization} while the pseudo-code for this step is presented in Algorithm \ref{alg:C-GOAT}. Note that the $\theta^m$ we use could be the default parameters of each $m\in M$ or the tuned parameter obtained in the Supervising step \ref{alg:S-GOAT}.
% As in standard for optimization problems, at each epoch of the tuning of the synthesizer models in Algorithm {\color{red} [REf]} and at each iteration of the Algorithm~\ref{alg:C-GOAT}, we optimize a loss function. Our optimization is done using the Bayesian approach that builds a probabilistic model around the parameters involved and then updates those parameters using the evaluation performance of the loss function. For the prior/posterior distribution over the objective function, we used the Parzen-Tree Estimator to find the location in the space of the parameters that maximize the expected improvement in the loss function. 

The meta-learning optimization formulation is given by:

\begin{flalign}
    \begin{aligned}
  \alpha ^ * =  \argmin_{\alpha \in  \mathcal{R} ^ {|M|}} \text{\quad} & \mathcal{L}(Y_{val}, \hat{Y}_{val}) \\
  \text{s.t \quad} & (X_{syn} ^ m, Y_{syn} ^ m) = S_{m}([ \alpha_m  N], \theta_m) \text{ } \forall m \in M \\
  & X_{syn} = [(X_{syn} ^ m) ^ T : m \in M] ^ T \\
  & Y_{syn} = [Y_{syn} ^ m : m \in M] \\
  & \hat{\mu} = f(Y_{syn}, X_{syn}) \\
  &\hat{Y}_{val} := \hat{\mu}(X_{val}) 
    \end{aligned}
    \label{eq:meta-learning-optimization}
 \end{flalign}

% \begin{flalign}
% \begin{aligned}
%   \min \text{\quad} & \mathcal{L}(Y_{val}, \hat{Y}_{val}) \\
% \text{s.t \quad} & \hat{Y}_{val} := \hat{\mu}_{\alpha ^*}(X_{val})
% \end{aligned}
% \label{eq:meta-learning-outer-optimization}
%  \end{flalign}

%  st 
 
% \begin{flalign}
% \begin{aligned}
% \alpha ^ * = \argmin_{\alpha} J(\alpha, D_{train})
% \end{aligned}
% \label{eq:meta-learning-inner-optimization}
%  \end{flalign}

where $\mathcal{L}$ refers to the loss function of the downstream task on the validation set $D_{val}$, which we use to evaluate the quality of the $\alpha's$ generated by the Bayesian optimization at each iteration. This involves evaluating the downstream task performance on the combined synthetic data generated using different methods as highlighted in Algorithm~\ref{alg:C-GOAT}.

\begin{algorithm} [tbh]
\caption{\textbf{C}omposing Step -   \textbf{G}enerative \textbf{O}ptimization \textbf{A}pproach for \textbf{T}abular data (C-GOAT)}\label{alg:C-GOAT}
\textbf{Input:} $D_{real}$, $\theta^m$ $\forall m \in M$\;
Create partition $\{D_{train}, D_{val}, D_{test}\}=\{(X_{train}, Y_{train}),(X_{val}, Y_{val}), (X_{test}, Y_{test})$\} \\
Initialize $\bm{\alpha}=\{\alpha_m\}_{m\in M}$\\
\For{$k = 1, \dots, K$}{
    Sample $D_m^k=S_m([\alpha_m^k N];\theta^m) \quad \forall m \in M$\\
    Create $D_{syn}^k=
    \begin{pmatrix}
    D_{GC}^k\\
  \hline
    \vdots\\
  \hline
  D_{TVAE}^k\\
    \end{pmatrix}=
    \begin{pmatrix}
    X_{GC}^k &\rvline& Y_{GC}^k\\
  \hline
    \vdots& \rvline &\vdots\\
  \hline
  X_{TVAE}^k &\rvline& Y_{TVAE}^k\\
    \end{pmatrix}$\\
    Train $\hat{\mu}^k=f(Y^k_{syn}\sim X^k_m)$ \\
    Compute $\hat{Y}_{val}^k=\hat{\mu}^k(X_{val})$\\
    Compute $l^k=\mathcal{L}(\hat{Y}_{val}^k,Y_{val})$\\
    Suggest $\bm{\alpha}^{k+1}$ using Bayesian optimization approach based on $\{\bm{\alpha}^{0},...,\bm{\alpha}^{k}\}$ and $\{l^0,...,l^k\}$\\
}
\textbf{return} $D_{syn}^{k^*}$ where $k^*=\argmin_{k} l^k$
\end{algorithm}

\subsection{Evaluation}

When generating synthetic data, one common concern is assessing the quality of the generated data. To evaluate synthetic generation models for tabular data, various benchmarking approaches are available, allowing flexibility in adapting the loss function to suit the specific objectives of synthetic data generation.

To evaluate the accuracy of preserving individual attributes and attribute pairs in synthetic data, the KS-Test and CS-Test are valuable tools. The KS-Test compares continuous column distributions using the empirical CDF \cite{fasano1987multidimensional}, while the CS-Test compares discrete column distributions using the Chi-Squared test \cite{patki2016synthetic}. Additionally, fidelity can be assessed by building a machine learning classifier to differentiate between real and synthetic data \cite{patki2016synthetic}.

While evaluating the distribution of synthetic and real data is crucial, we must also address privacy protection at an individual level. Like many machine learning models, synthetic generative approaches are susceptible to privacy attacks \cite{sun2021adversarial}, including Membership Inference Attacks (MIA) \cite{shokri2017membership}, Reconstruction attacks \cite{narayanan2006break}, and Property inference attacks \cite{lin2023summary}. Addressing these privacy vulnerabilities is crucial to preserving the utility and integrity of synthetic data.

In this paper, we aim to evaluate our synthetic generative models through the lens of the downstream classification model's accuracy, which serves as a robust metric to assess the models' overall performance.

\section{Data}\label{sec:data-sets}
% Saheed
\paragraph{\textbf{Adult}}
This data set is a sample from the US Census Bureau Database that contains the census result of the year 1994\footnote{Data available on the UCI platform at \url{https://archive.ics.uci.edu/data set/2/adult}}. This data set includes 48,842 records and 14 attributes. Each record contains the following features such as age, gender, education, relationship, occupation, race, and native country of a representative individual in the census record. These attributes are a mixture of numerical, ordinal, and categorical data types. The data set has a binary target label which indicates whether the income of an individual is less or greater than fifty thousand dollars. Therefore, the data set has a classification task which is to predict if a person makes over 50K a year based on the census attributes

\paragraph{\textbf{Credit Card Fraud}}
To showcase the usefulness of synthetic tabular data, we use the credit card fraud data set\footnote{Data available on the Kaggle platform at \url{https://www.kaggle.com/data sets/mlg-ulb/creditcardfraud}}. This data set contains transactions collected in the span of two days made by credit cards by European cardholders for the month of September 2013. From an analysis, it can be observed that the data set is highly imbalanced containing 492 frauds out of 28,4807 total transactions. The positive class of fraud accounts for 0.172\% of all transactions. The credit card fraud data set contains only numerical input variables with 31 features. With respect to confidentiality and privacy, 28 of the features - V1 to V28 are principal components obtained by the means of PCA. 'Time', 'Amount', and 'Class' are the only features not to be transformed with PCA. The feature 'Time' contains the seconds elapsed between each transaction and the first transaction in the data set. The feature 'Class' is the target variable which takes the value of 0 for cases of no fraud and 1 for cases of fraud. The feature 'Amount' is the transaction amount. Given the class imbalance ratio of the credit fraud data set, we processed the data set by oversampling the minority class with random undersampling of the majority class, leading to a more balanced data set. This involved duplicating examples in the minority class in order to reach an equal balance between the minority and majority class. This process will reduce the number of data points available. This technique is called Synthetic Minority Oversampling Technique (SMOTE)\cite{chawla2002smote}. Applying this technique will lead us to have two separate data sets, an original imbalanced credit data set and a new balanced credit data set.

This data set will be helpful in the context of fraud detection for machine learning utility. We can answer whether synthetic data generation can help with downstream tasks in the fraud management process. The utility of models trained on fraud data sets allows us to measure the effectiveness of detecting and predicting potential fraudulent operations. This provides guidance to fraud practitioners interested in utility using synthetic data to train fraud detection models.

\section{Experimental Results}\label{sec:experimental-results}

We evaluate our approach on three diverse data sets discussed in the previous section: the adult  data set\footnotemark[1], the balanced credit card data set\footnote{The credit card data set with oversampling the minority class as mentioned in Section~\ref{sec:data-sets}.}, and the imbalanced credit card data set\footnotemark[2]. We chose these data sets as they are widely utilized in previous works for evaluating tabular synthetic data generation methods. The adult data set contains both numerical and categorical variables allowing us to showcase the applicability of our approach in generating different types of data. 

Furthermore, by selecting both balanced and imbalanced data sets, we can demonstrate the robustness of our approach across various data distributions. The data set's descriptions are summarized in Table \ref{table:description-data sets}. For the adult data set, we utilized all available records, totaling $48.842$K. However, for the credit card data set, we sampled $50$K records from the available $28.407$K records similar to \cite{zhao2022ctab}. Through this evaluation, we gain valuable insights into the generalizability and performance of our approach, enhancing its credibility as a powerful tool for generating high-quality synthetic data across a diverse range of scenarios.

\begin{table*}[ht!]
  \caption{Description of data sets}
  \label{table:description-data sets}
  \centering
  % \footnotesize
  \tiny
  \begin{tabular}{llllllll}
    \toprule
    Data set  & Label & Observation & Continuous & Binary & Multi-class & Label = 0 & Label = 1 \\
    \midrule
    Adult \footnotemark[1] & 'income' & 48,842 & 6 & 2 & 7 & 76.07\% & 23.93\% \\
    Credit Balanced \footnotemark[3] & 'Class' & 50,000  & 30  &  1 & 0 & 66.70\%\% & 33.3\%\\
    Credit Imbalanced \footnotemark[2] & 'Class' & 50,000 & 30  &  1 & 0 & 99.82\% & 0.18\%\\
    \bottomrule
  \end{tabular}
\end{table*}

\begin{tiny}

\begin{table}[tb!]
\caption{Description of synthetic data sets generated using each model.}
\label{table:description-data sets-synthetic-tuned}
\centering
\begin{tabular}{llllll}
\toprule
 &  & \multicolumn{2}{c}{Untuned} & \multicolumn{2}{c}{Tuned} \\
Data set  & Method  & Label = 0 & Label =1    & Label = 0    & Label =1   \\
\midrule
\multirow{4}{*}{Adult}  & Gaussian Copula \cite{SDV} & 80.40\%   & 19.60\% & 82.14\%  & 17.9\% \\
 &  CTGAN \cite{SDV}  & 83.66\%   & 16.34\% & 77.15\%  & 22.9\% \\
 & CopulaGAN \cite{SDV} & 74.96\%   & 25.04\% & 74.96\%  & 25.0\% \\
 & TVAE  \cite{SDV}  & 76.44\%   & 23.56\% & 77.63\%  & 22.4\% \\
 & Our method    & 78.47\%  & 21.53\%  & 77.41\% & 22.59\% \\
  \midrule
\multirow{4}{*}{Adult Transformed \tablefootnote{This is the original adult data set with target encoder transformation for the categorical features.}} & Gaussian Copula & 94.29\%   & 5.71\%  & 80.36\%  & 19.6\% \\
 & CTGAN  & 68.63\%   & 31.37\% & 77.45\%  & 22.6\% \\
 & CopulaGAN & 74.85\%   & 25.15\% & 74.85\%  & 25.2\% \\
 & TVAE    & 78.70\%   & 21.30\% & 76.32\%  & 23.7\% \\
 & Our method    & 70.19\%  & 29.81\%  & 72.93\% &27.07\% \\
 \midrule
\multirow{4}{*}{Credit Balanced}   & Gaussian Copula   & 70.07\%   & 29.93\% & 63.66\%  & 36.3\% \\
 & CTGAN  & 68.33\%   & 31.67\% & 63.73\%  & 36.3\% \\
 & CopulaGAN & 91.97\%   & 8.03\%  & 91.97\%  & 8.0\%  \\
 & TVAE    & 63.90\%   & 36.10\% & 63.90\%  & 36.1\% \\
 & Our method    & 63.90\%  & 36.10\%  & 63.90\% & 36.1\% \\
  \midrule
\multirow{4}{*}{Credit Imbalanced} & Gaussian Copula   & 0.00\% & 100.00\%    & 0.00\%   & 100.0\%    \\
 & CTGAN  & 100.00\%  & 0.00\%  & 100.00\% & 0.0\%  \\
 & CopulaGAN & 99.82\%   & 0.18\%  & 99.82\%  & 0.2\%  \\
 & TVAE    & 100.00\%  & 0.00\%  & 100.00\% & 0.0\% \\
  & Our method    & 99.65\%  & 0.35\%  & 99.65\% &0.35\% \\
 \bottomrule
\end{tabular}

\end{table}
\end{tiny}
Our method is implemented as an open-source Python package that will be available on GitHub. The implementation utilizes four generative methods, namely Gaussian Copula, CTGAN, Copula GAN, and TVAE, available from the SDV \cite{SDV} python package. For the downstream task evaluation, we utilize the XGBoost classifier python package \cite{chen2016xgboost}, and for the Bayesian optimization, we use the hyperopt python package \cite{bergstra2013making}.

All experiments were conducted using Python 3.10. The code repository provides comprehensive instructions for replicating the experiments and includes detailed result tables. For further insights into the generated synthetic data sets, a summary is provided in Table~\ref{table:description-data sets-synthetic-tuned} for one experiment.

Each experiment was repeated 10 times and for these experiments, 70\% of the real data was used for training, 20\% for validation, and the remaining 10\% for testing. For the untuned setup, we set $K = 350$ in Algorithm~\ref{alg:S-GOAT}, while for the tuned setup, we used $K = 150$ in Algorithm~\ref{alg:C-GOAT}. For Algorithm~\ref{alg:C-GOAT}, we generate the alphas using the uniform distribution $Uniform(0, 1)$ and then we scale those alphas to add up to 1 as follows:
\begin{equation*}
    \alpha_m = \frac{\alpha_{m} }{\sum_{j \in M} \alpha_{j} } \quad \forall m \in M
\end{equation*}

For the first iteration, instead of randomly generating the alphas, we use a warm start. The warm start also addresses a weakness in the Bayesian optimization using the Tree-Structured Parzen-Estimator, as in the majority of cases, it fails to converge to the optimal solution if it lies on one of the corner points. By "corner points" here, we mean that the optimal solution only considers the best method and neglects the others, which can be mathematically defined as $\alpha = 1$ for the best model and $\alpha = 0$ for the other methods. This decision is based on the evaluation metric for the validation data set, which, in our implementation, is the AUC score. Therefore, our approach considers five initial starting points. Four of them represent the corner points that correspond to each model $m \in M$. For example, the initial starting point that only represents the Gaussian Copula model will be given by $[1.0, 0.0, 0.0, 0.0]$. The last initial starting point is initialized based on the AUC validation of each individual model:
\begin{equation*}
    \alpha_m = \frac{auc_{val}^{m} -  auc^*_{val}}{\sum_{j \in M} (auc_{val} ^ {j} -auc^*_{val})} \quad \forall m \in M
\end{equation*}

where $auc^*_{val} = min_{i\in M} auc_{val}^{i}$. 

The decision for which initial point to pick depends on the point that gives the best validation AUC score for the first iteration. This initialization scheme ensures that in case our optimal solution lies on one of the corner points, we will converge to it. Additionally, if the optimal solution represents a mixture of each individual model, we will also be able to capture it, as the algorithm will generate alpha vectors that continuously improve the loss function. 

However, to prevent the chance of overfitting, we implemented early stopping by adding a condition to stop the algorithm once the AUC score on the validation set doesn't improve for the last $k$ iterations. For algorithm \ref{alg:S-GOAT}, we used $k = 10$, and for algorithm \ref{alg:C-GOAT} we used $k = 15$.

Our results include fitting all the individual models from the SDV package \cite{SDV} without any hyper-parameter tuning as well as tuning these models as mentioned in Algorithm \ref{alg:S-GOAT} . For CTAB-GAN+, the details of the experiment are mentioned in Subsection~\ref{subsec:ctab-gan-plus}. We also reported the results of our method using both tuned and untuned setups where for the untuned setup we only use Algorithm~\ref{alg:C-GOAT} with untuned models from the SDV package \cite{SDV} while for the tuned setup we use both Algorithm~\ref{alg:S-GOAT}
 and Algorithm~\ref{alg:C-GOAT}.
 
\begin{table}[tbh]
    \centering
    \caption{Average test AUC for the XGBoost baseline model fitted only on real data for each data set for 10 experiments.}
    \label{table:baseline-model-results}
    \begin{tabular}{llll}
  \toprule
   Method & Adult &  Credit Balanced  & Credit Imbalanced\\
   \midrule
   XGBoost \cite{chen2016xgboost}  & 90.51\% & 99.99\% & 96.02\% \\
   \bottomrule
    \end{tabular}
    \label{tabel:baseline-model-results}
\end{table}

\subsection{Performance Evaluation}
To optimize the loss function, we aim to maximize the AUC score for the downstream classification task. This is achieved by training an XGBoost classifier \cite{chen2016xgboost} on the training data set and subsequently evaluating its performance on a separate validation data set. To ensure a fair comparison between the different methods, the XGBoost classifier is utilized with its default parameters.  By focusing on the maximization of the AUC score in the downstream task, we can accurately evaluate the synthetic data's quality. Our primary objective is to generate data that closely resembles the real data. Therefore, by emphasizing the AUC score, we ensure that the synthetic data is as representative as possible, enabling it to capture essential characteristics and patterns present in the real data set.

\subsection{Baseline model}\label{sec:baseline-model}
To comprehensively evaluate the effectiveness and improvements of our method, as well as the quality of the generated data compared to the original real data set, we fitted XGBoost on the original data and assessed the model's performance in terms of AUC. By comparing our results against the baseline XGBoost model fitted on real data, we gain valuable insights into the efficiency of our approach and the similarity between the generated synthetic data and the real data.

%Fadi
\subsection{CTAB-GAN+}\label{subsec:ctab-gan-plus}
Given that the primary criteria for evaluating our approach rely on downstream losses, we conduct a thorough comparison against CTAB-GAN+ \cite{zhao2022ctab}.
% , a cutting-edge method for generating tabular synthetic data \cite{zhao2022ctab}. 
CTAB-GAN+ stands out as a novel conditional tabular GAN, surpassing existing state-of-the-art approaches by incorporating downstream losses into conditional GANs. This innovation results in higher utility synthetic data that proves beneficial in both classification and regression domains. 
% Notably, CTAB-GAN+ introduces novel encoders specifically designed to handle mixed continuous-categorical variables and variables with unbalanced or skewed data. 
The model introduces several other major improvements over existing methods \cite{zhao2022ctab}. As we compare our approach to this state-of-the-art alternative, we aim to showcase the strengths and competitive advantages of our synthetic data generation technique in practical scenarios. 

We focus on the default version of CTAB-GAN+ without any fine-tuning. The decision is driven by time constraints, as tuning the model requires a significant amount of time compared to the models present in SDV. By considering the default CTAB-GAN+, we can still gain valuable insights and effectively evaluate the relative strengths of our method without the need for extensive fine-tuning efforts. Table \ref{table:results-all-models} summarizes the results.

\subsection{Adult data set}
We evaluated the efficiency of our method on the adult data set\footnotemark[1], focusing on its performance with categorical data. For encoding the categorical features, we employed two distinct approaches. The first approach relied on the implicit handling by the SDV python package \cite{SDV} during the fitting of synthesizers on the real data. This implicit handling is implemented using a label encoder. The second approach involved implementing a target encoder, which outperforms traditional encoding schemes, especially for categorical features with high cardinality in the categorical columns \cite{micci2001preprocessing}. We referred to the transformed data, utilizing the target encoder, as the 'Adult Transformed' data set.

Subsequently, we compared the performance of our approach against each individual synthesizer model $m \in M$ using two setups: untuned models and tuned models, as described in section~\ref{sec:supervising-synthesizers}. For the 'income' column, we mapped rows with values '<=50k' to 0 and rows with values '>50k' to 1. The results of the comparison are summarized in Table ~\ref{table:results-all-models}.

\subsection{Balanced credit card data set}

We further demonstrate the effectiveness of our approach on the credit card data set\footnotemark[2], which exclusively comprises numerical features. Initially, the credit card data set exhibited an imbalanced distribution, as indicated in Table~\ref{table:description-data sets}. However, as detailed in Section~\ref{sec:data-sets}, we preprocessed the data set using a random undersampling of the majority class. This resulted in a more balanced data set, facilitating a fairer evaluation.

Similarly to our approach for the adult data set, we conduct a thorough comparison of our method's performance against that of each individual synthesizer model $m \in M$, employing two setups: untuned models and tuned models, as described in Section~\ref{sec:supervising-synthesizers}. The results of this comparison are summarized in Table ~\ref{table:results-all-models}.

\subsection{Imbalanced credit card data set}

Considering real-life scenarios often involve highly imbalanced data, we evaluated the performance of our approach on an imbalanced credit card data set. Initially, we ran our approach using four synthesizer methods in both 'tuned' and 'untuned' setups. However, we observed weaknesses in handling highly imbalanced data sets, as these methods struggled to generate data from both classes effectively. For instance, TVAE only generated data from the majority class, while CopulaGAN unexpectedly generated data solely from the minority class. To address this issue, we implemented conditional sampling, with only Gaussian Copula successfully generating data resembling the original data set. The results of this comparison are summarized in Table ~\ref{table:results-all-models}.
% Despite this unusual data generation behavior, our approach outperformed individual methods, successfully recreating the original data distribution with a high AUC score. The comparison of results is presented in Table~\ref{table:results-all-models}. 
% Through these efforts, we achieved significant improvements in handling imbalanced data sets and generating data that closely matches the characteristics of the real data set.

% \usepackage{multirow}
\begin{table*}[]
\caption{Contribution of each individual model ($\alpha$) for the final synthetic data generated.}
\label{table:contribution-individual-models}
\begin{tabular}{llll}
\toprule
Data set    & Method & Weights (untuned setup) & Weights (tuned setup) \\
\midrule
\multirow{4}{*}{Adult}    & Gaussian Copula & 20.90\% & 17.79\%   \\
 & CTGAN  & 29.50\% & 41.83\%   \\
 & CopulaGAN   & 19.07\% & 14.26\%   \\
 & TVAE   & 30.53\% & 26.12\%   \\
 \midrule
\multirow{4}{*}{Adult Transformed} & Gaussian Copula & 26.69\% & 19.44\%   \\
 & CTGAN  & 32.98\% & 42.72\%   \\
 & CopulaGAN   & 8.22\%  & 11.51\%   \\
 & TVAE   & 32.11\% & 26.33\%   \\
 \midrule
\multirow{4}{*}{Credit Balanced}   & Gaussian Copula & 0\% & 0\%   \\
 & CTGAN  & 0\%  & 0\%    \\
 & CopulaGAN   & 0\%  & 0\%    \\
 & TVAE   & 100\% & 100\%   \\
 \midrule

\multirow{4}{*}{Credit Imbalanced} & Gaussian Copula & 99.33\% & 99.33\%   \\
 & CTGAN  & 0.22\% & 0.22\%   \\
 & CopulaGAN   & 0.22\%  & 0.22\%    \\
 & TVAE   & 0.22\% & 0.22\%  \\
 \bottomrule
\end{tabular}
\end{table*}

\begin{table}[thb!]
\caption{Average, standard deviation, and one-sided paired t-test for the downstream test AUC score, using XGBoost fitted on the generated data by each method, on 10 experiments.}
\label{table:results-all-models}
% \footnotesize
\tiny
\centering
\begin{tabular}{llllllllll}
\toprule
     &  & \multicolumn{4}{c}{Untuned}  & \multicolumn{4}{c}{Tuned}  \\
Data set   & Method     & average & std & test statistic & p-value & average & std & test statistic & p-value \\
\midrule
\multicolumn{1}{l}{\multirow{6}{*}{Adult}} & Gaussian Copula & 76.43\% & 0.04    & 10.07 & 0   & 76.43\% & 0.04    & 10.24 & 0   \\
\multicolumn{1}{c}{}  & CTGAN  & 83.12\% & 0.02    & 9.39  & 0   & 84.28\% & 0.01    & 7.84  & 0   \\
\multicolumn{1}{c}{}  & CopulaGAN  & 80.18\% & 0.03    & 7.59  & 0   & 79.58\% & 0.03    & 9.12  & 0   \\
\multicolumn{1}{c}{}  & TVAE   & 82.59\% & 0.01    & 14.25 & 0   & 82.92\% & 0.01    & 10.97 & 0   \\
\multicolumn{1}{c}{}  & CTAB-GAN+    & 83.25\% & 0.03    & 4.44  & 0.0002  & -   & -   & - & -   \\
\multicolumn{1}{c}{}  & Our method & \textbf{87.87}\% & 0.00    & - & -   & \textbf{88.12}\% & 0.01    & - & -   \\
\midrule
\multirow{6}{*}{Adult Transformed}     & Gaussian Copula & 76.06\% & 0.03    & 9.60  & 0   & 76.06\% & 0.03    & 9.95  & 0   \\
   & CTGAN  & 77.21\% & 0.04    & 7.20  & 0   & 78.62\% & 0.05    & 5.23  & 0   \\
   & CopulaGAN  & 71.24\% & 0.06    & 8.37  & 0   & 73.02\% & 0.05    & 9.56  & 0   \\
   & TVAE   & 81.86\% & 0.02    & 6.23  & 0   & 81.07\% & 0.03    & 5.49  & 0   \\
   & CTAB-GAN+    & 81.81\% & 0.03    & 4.76  & 0.0001  & -   & -   & - & -   \\
   & Our method & \textbf{86.67}\% & 0.01    & - & -   & \textbf{86.97}\% & 0.01    & - & -   \\
   \midrule
\multirow{6}{*}{Credit Balanced} & Gaussian Copula & 94.45\% & 0.01    & 14.10 & 0   & 94.45\% & 0.01    & 14.31 & 0   \\
   & CTGAN  & 95.34\% & 0.01    & 16.21 & 0   & 95.93\% & 0.01    & 13.15 & 0   \\
   & CopulaGAN  & 95.50\% & 0.01    & 14.18 & 0   & 96.41\% & 0.01    & 7.80  & 0   \\
   & TVAE   & 98.52\% & 0.00    & 0.00  & 0.5 & 98.48\% & 0.00    & 0.00  & 0.5 \\
   & CTAB-GAN+    & 98.04\% & 0.00    & 2.74  & 0.0068  & -   & -   & - & -   \\
   & Our method & \textbf{98.52}\% & 0.00    & - & -   & \textbf{98.48}\% & 0.00    & - & -   \\
   \midrule
\multirow{6}{*}{Credit Imbalanced}     & Gaussian Copula & 95.32\% & 0.07    & -0.23 & 0.5886  & 95.32\% & 0.07    & 0.47  & 0.3215  \\
   & CTGAN  & 50.00\% & 0.00    & 19.17 & 0   & 50.00\% & 0.00    & 28.47 & 0   \\
   & CopulaGAN  & 50.00\% & 0.00    & 19.17 & 0   & 50.00\% & 0.00    & 28.47 & 0   \\
   & TVAE   & 50.00\% & 0.00    & 19.17 & 0   & 50.00\% & 0.00    & 28.47 & 0   \\
   & CTAB-GAN+    & \textbf{96.70}\% & 0.05    & -0.74 & 0.7667  & -   & -   & - & -   \\
   & Our method & 94.60\% & 0.07    & - & -   & \textbf{96.59}\% & 0.05    & - & -  \\
   \bottomrule
\end{tabular}
\end{table}

\subsection{Results Analysis}\label{sec:results-analysis}
\begin{figure}[h]
\begin{center}
  \includegraphics[scale=0.35]{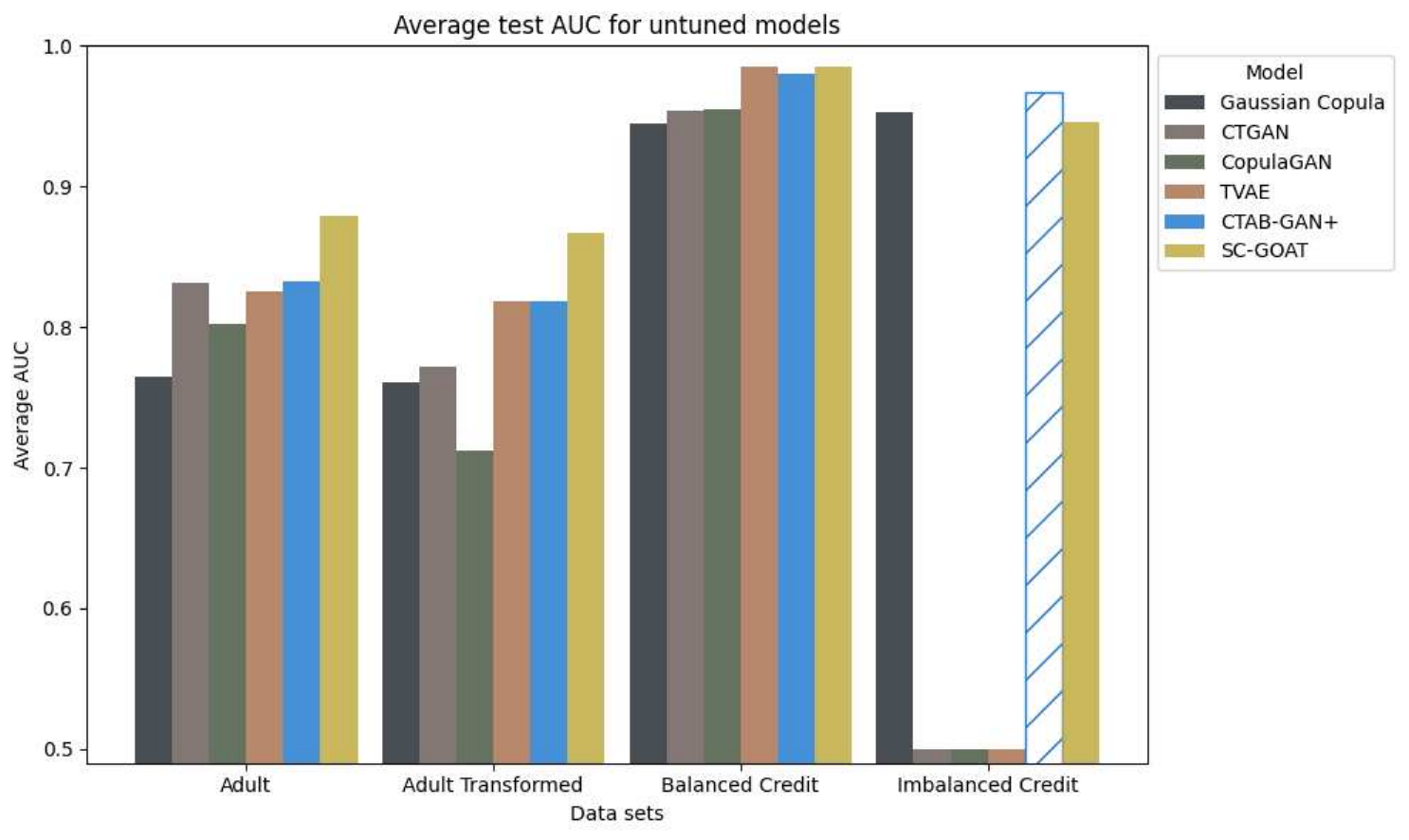}
  \label{fig:test-untuned}
\end{center}
% \caption{Average downstream test AUC score for 10 experiments using XGBoost fitted on the generated data by each model in the untuned setup.}
\caption{Average downstream test AUC score for 10 experiments using XGBoost fitted on the generated data by each model in the untuned setup\protect\footnotemark[5].}
\label{fig:results-test}
\end{figure}

To simplify the comparison between our approach and the other different methods, and to demonstrate the applicability of our method in various scenarios, we present the results averaged across all the experiments for all the methods on the test data in Figure \ref{fig:results-test}. Here we only show the plot comparison for the untuned setup since we didn't tune the CTAB-GAN+.
% as the model takes a significant amount of time to be tuned. 

It can be observed that our method outperforms all the other approaches in most cases. Only for the imbalanced credit card data set, CTAB-GAN+ performs better than our approach. This is related to the way CTAB-GAN+ handles imbalanced data, as they implement a training-by-sampling strategy. The idea behind their approach is to resample classes, giving higher chances to minority classes to train the model \cite{zhao2022ctab}. This approach is somehow similar to making the data balanced as we did in Subsection~\ref{sec:data-sets} to handle the credit card data, and this similarity can be observed clearly on the balanced data set, as both approaches perform the same (refer to Figure~\ref{fig:results-test}). This means that comparing our approach with CTAB-GAN+ on imbalanced data isn't fair due to the extra step they implement. Even without implementing any additional upsampling, our approach performed very close to CTAB-GAN+ on the imbalanced data (refer to Table~\ref{table:results-all-models})\footnote{In Figure \ref{fig:results-test}, we illustrate CTAB-GAN+ for the imbalanced data set using a partial filled color to emphasize the distinction between their approach and ours in handling imbalanced data, as described in Section~\ref{sec:results-analysis}.}.
% in the untuned setup; however, tuning results in a significant improvement in our method, matching the performance of CTAB-GAN+.

%FH
A major part of our work is providing a metric to understand the most suitable synthesizer among multiple synthesizers for a specific objective. To demonstrate that, we present in Table~\ref{table:contribution-individual-models} the values of alphas for each model from one of the experiments. We can clearly see from this table that the alpha weights are linked to the performance of each individual model, and as expected, the model with better performance has a higher $\alpha$ weight compared to the other models. In the balanced credit card data, the corner solution emerged as the winner, with all weights assigned to TVAE owing to its superior validation AUC before early stopping. For the imbalanced credit card data, the warm-started weight achieved the highest downstream AUC in most cases. Lastly, in the adult dataset, a mixture of the four methods was used, and Algorithm~\ref{alg:C-GOAT} learned the optimal weight based on the downstream validation AUC.
% However, the advantage of our method lies within the optimization part that doesn't limit the sampling to only the best model, as the combination of the data from each separate model is the key to improving the quality of the generated data. 

Comparing against the baseline model fitted only on real data as explained in Subsection~\ref{sec:baseline-model}, we aim to show that the generated synthetic data have a similar downstream performance. Comparing the values from Tables~\ref{table:baseline-model-results} and ~\ref{table:results-all-models} shows that the AUC test scores for the XGBoost fitted only on real data are very close to the AUC test scores for the XGBoost fitted only on synthetic data. This validates our point that the distribution of the data generated by our approach can perform a high-quality downstream task.

We could observe that tuning the neural network-related hyperparameters through Algorithm \ref{alg:S-GOAT} did not lead to a significant performance boost. This finding raises two possibilities: we may need to expand the hyperparameter grid and explore the space more extensively, or the three methods we tuned inherently exhibit robustness to hyperparameters. We recognize this as an open discussion for further investigation. If the latter holds, practitioners could potentially skip the costly hyperparameter tuning step, and use out untuned set-up leading to a more efficient synthetic data creation process.

\section{Conclusion and future work}
Our approach has shown great promise, consistently outperforming the majority of previous methods in terms of the downstream metric. For future work, one potential direction is to evaluate the performance of SC-GOAT concerning privacy or fidelity aspects. Moreover, this approach could be further explored for data augmentation purposes, aiming to surpass the downstream metric achieved with real data. Such investigations could provide valuable insights and advancements in the field.

\textbf{Disclaimer} This paper was prepared for informational purposes by the Artificial Intelligence Research group of JPMorgan Chase \& Co and its affiliates (“J.P. Morgan”), and is not a product of the Research Department of J.P. Morgan. J.P. Morgan makes no representation and warranty whatsoever and disclaims all liability, for the completeness, accuracy or reliability of the information contained herein. This document is not intended as investment research or investment advice, or a recommendation, offer or solicitation for the purchase or sale of any security, financial instrument, financial product or service, or to be used in any way for evaluating the merits of participating in any transaction, and shall not constitute a solicitation under any jurisdiction or to any person, if such solicitation under such jurisdiction or to such person would be unlawful.
%%%%%%%%%%%%%%%%%%%%%%%%%%%%%%%%%%%%%%%%%%%%%%%%%%%%%%%%%%%%

\newpage

\bibliographystyle{unsrtnat}
\bibliography{sample}

\end{document}